\documentclass[10pt, conference, compsocconf]{IEEEtran}

\usepackage{bibspacing}
\usepackage{algorithm,algpseudocode}
\usepackage{subcaption}
\usepackage{adjustbox}
\usepackage{amsfonts}
\usepackage{amsmath}
\usepackage{array}
\usepackage[colorinlistoftodos,prependcaption,textsize=tiny]{todonotes}
\usepackage{bm}
\usepackage{bbm}
\usepackage{booktabs}
\usepackage{braket}

\usepackage{cite}
\usepackage{color}

\usepackage{dsfont}

\usepackage{eqparbox}

\usepackage{float}

\usepackage{graphicx}

\usepackage{mathtools}
\usepackage{mdframed}
\usepackage{mdwmath}
\usepackage{mdwtab}

\usepackage{paralist}

\usepackage{setspace}

\usepackage{tikz,pgfplots,pgfplotstable}
\usetikzlibrary{pgfplots.groupplots}
\usetikzlibrary{arrows, backgrounds, calc, hobby, positioning}
\usetikzlibrary{shapes.geometric}
\pgfplotsset{compat=1.9}
\usepackage{url}

\usepackage{xspace}

\newmdtheoremenv{obs}{Observation}

\numberwithin{subcase}{case}

\long\def\/*#1*/{}

\DeclareMathOperator*{\argmin}{argmin}

\algnewcommand{\LineComment}[1]{\Statex \(\triangleright\) #1}

\begin{document}
\title{S-Isomap++: Multi Manifold Learning from Streaming Data}

\author{
  \IEEEauthorblockN{Suchismit Mahapatra}
\IEEEauthorblockA{Computer Science and Engineering\\
State University of New York at Buffalo\\
suchismi@buffalo.edu}
\and
  \IEEEauthorblockN{Varun Chandola}
\IEEEauthorblockA{Computer Science and Engineering\\
State University of New York at Buffalo\\
chandola@buffalo.edu}}

\maketitle
\IEEEpeerreviewmaketitle

\begin{abstract}
Manifold learning based methods have been widely used for non-linear dimensionality reduction (NLDR). However, in many practical settings, the need to process streaming data is a challenge for such methods, owing to the high computational complexity involved. Moreover, most methods operate under the assumption that the input data is sampled from a single manifold, embedded in a high dimensional space. We propose a method for streaming NLDR when the observed data is either sampled from multiple manifolds or irregularly sampled from a single manifold. We show that existing NLDR methods, such as Isomap, fail in such situations, primarily because they rely on smoothness and continuity of the underlying manifold, which is violated in the scenarios explored in this paper. However, the proposed algorithm is able to learn effectively in presence of multiple, and potentially intersecting, manifolds, while allowing for the input data to arrive as a massive stream. 
\end{abstract}


\section{Introduction}
\label{sec:introduction}
Ability to analyze massive streams of data is a valuable aspect of any modern data science pipeline. This is important in many contexts, such as high-performance high-fidelity numerical simulations~\cite{duan2013}, high-resolution scientific instrumentation (microscopes, DNA sequencers, etc.)~\cite{ruan2014}, and even {\em Internet of Things}~\cite{chen2015}, where a huge number of devices are currently connected to the Internet and feeding a variety of data streams. Such data sources typically monitor or measure complex system behaviors, using a large number of parameters. Dimensionality reduction methods~\cite{vandermaaten2009} are typically used to map the resulting high-dimensional data into a smaller, manageable space. If the data is assumed to lie on a hyperplane, linear dimensionality reduction methods such as Principal Component Analysis (PCA)~\cite{hotelling1933}, etc., maybe applied. However, in many settings, especially when dealing with complex scientific and natural phenomenon, the data might lie on a non-linear manifold, in which case, non-linear dimensionality reduction methods are more appropriate.

Non-linear dimensionality reduction(NLDR) comes at a cost; most existing NLDR methods have a computational complexity of $O(n^3)$, $n$ being the size of the data. The issue is further exacerbated when the data is streaming, where obtaining exact solution at every step of the stream is computationally infeasible. While adapatations of existing NLDR methods, such as Isomap~\cite{tenenbaum2000} and Local Linear Embedding (LLE)~\cite{roweis2000}, have been proposed for handling data streams~\cite{kouropteva2005,law2006}, such methods, which typically rely on incremental updates of the underlying solution, do not scale well to massive streams. In a recent work~\cite{schoeneman2017}, a two phase strategy has been proposed to adapt Isomap to streaming data. The algorithm, called {\em S-Isomap}, operates on the core principle that a small batch of data is necessary to {\em learn} the underlying small-dimensional manifold using an exact and computationally expensive, but data-bounded, learning method. The remainder of the stream may be {\em mapped} onto the learnt manifold using a relatively inexpensive mapping procedure.

\begin{figure*}[t]
\centering
\begin{subfigure}{0.3\textwidth} 
  \includegraphics[height=0.2\textheight]{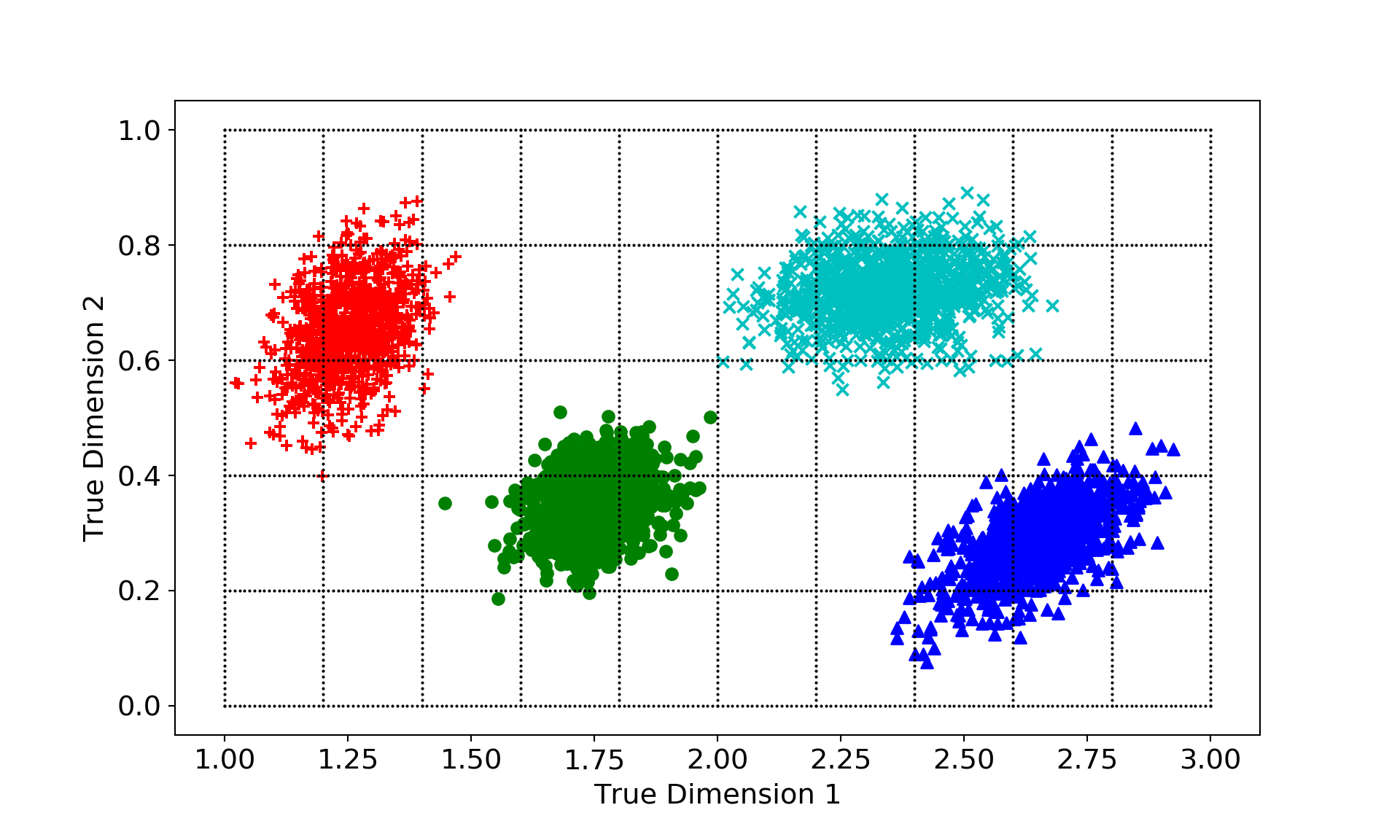} 
  \caption{Original Data in 2D}
  \label{fig:patchesa}
\end{subfigure}
\begin{subfigure}{0.35\textwidth} 
  	\includegraphics[height=0.2\textheight]{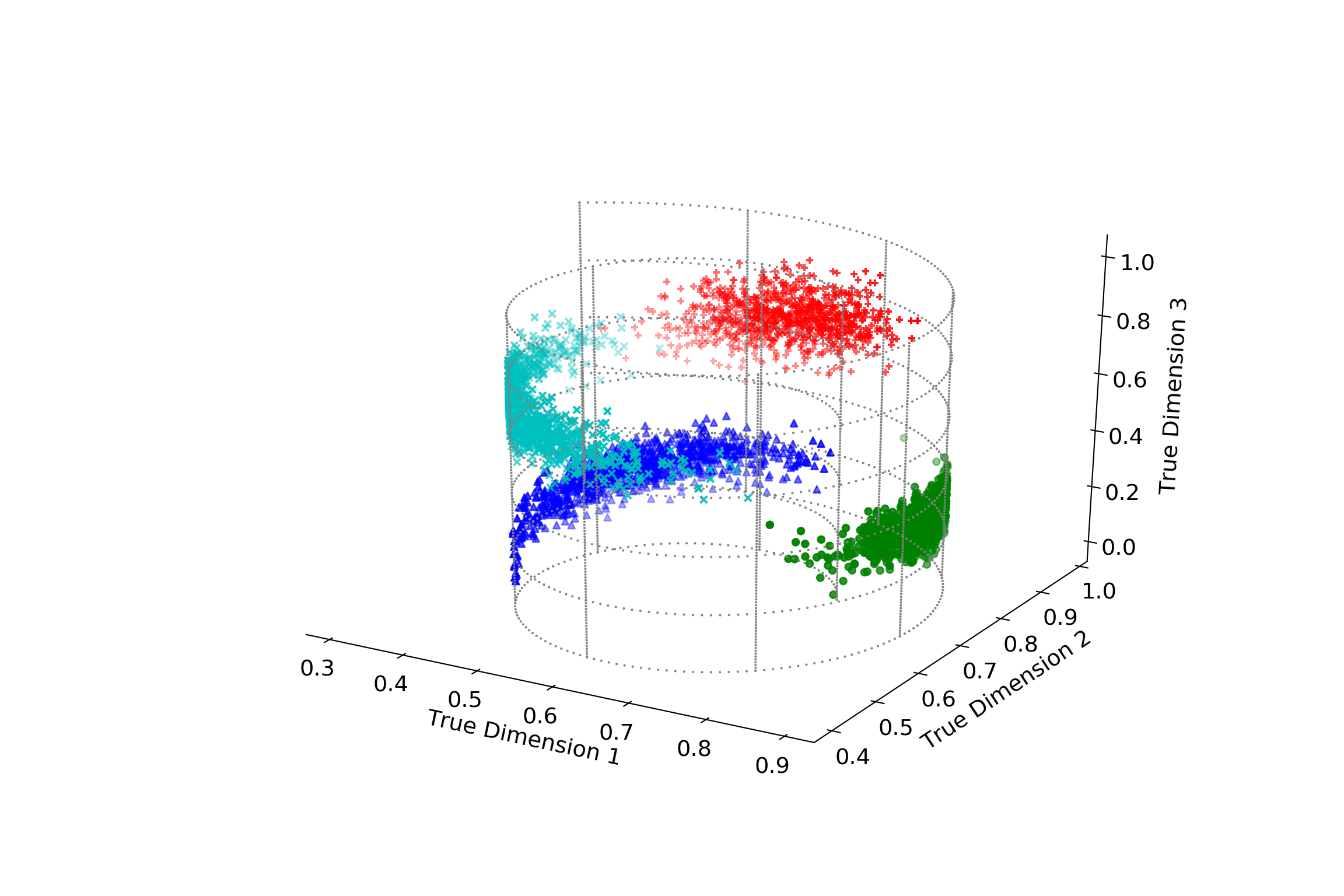}
  \caption{Embedded Data (input) in 3D}
    \label{fig:patchesb}
\end{subfigure}
\begin{subfigure}{0.3\textwidth}
\includegraphics[height=0.2\textheight]{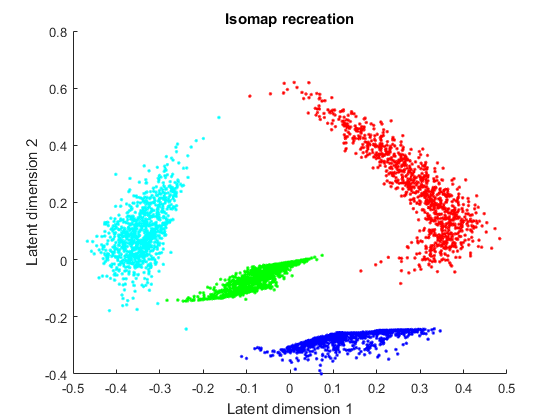} 
  \caption{Isomap Output}
    \label{fig:patchesc}
\end{subfigure}
\begin{subfigure}{0.3\textwidth}
\includegraphics[height=0.2\textheight]{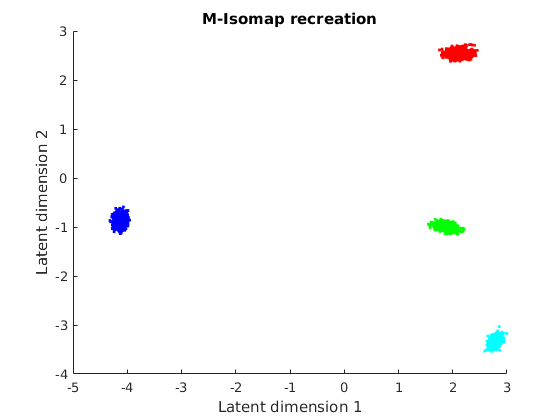}
  \caption{M-Isomap Output}
    \label{fig:patchesd}
\end{subfigure}
\begin{subfigure}{0.3\textwidth}
\includegraphics[height=0.2\textheight]{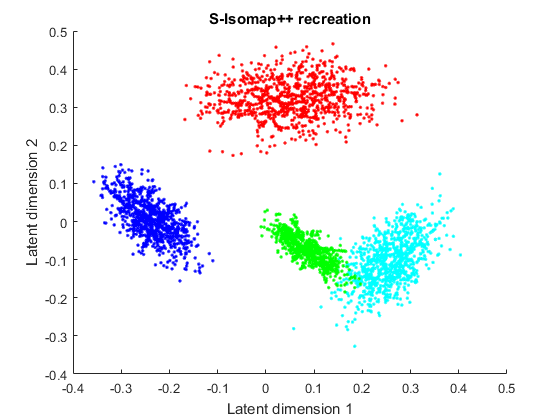}
  \caption{Proposed S-Isomap++ Output}
    \label{fig:patchese}
\end{subfigure}
\caption{Multi-manifold {\em patches} data set. The 2D samples in (a) are embedded into 3D in (b) via the Euler Isometric mapping technique~\cite{schoeneman2017}. The reduction to 2D is obtained using: (c). Isomap, (d). M-Isomap~\cite{fan2012}, and (e). the proposed S-Isomap++ algorithm.}
\label{fig:patches}
\end{figure*}
\begin{figure*}[t]
\centering
\begin{subfigure}{0.3\textwidth} 
  \includegraphics[height=0.2\textheight]{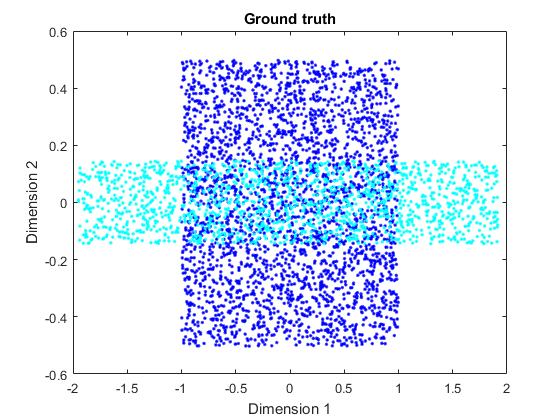} 
  \caption{Original Data in 2D}
  \label{fig:intersectinga}
\end{subfigure}
\begin{subfigure}{0.3\textwidth}  
  	\includegraphics[height=0.2\textheight]{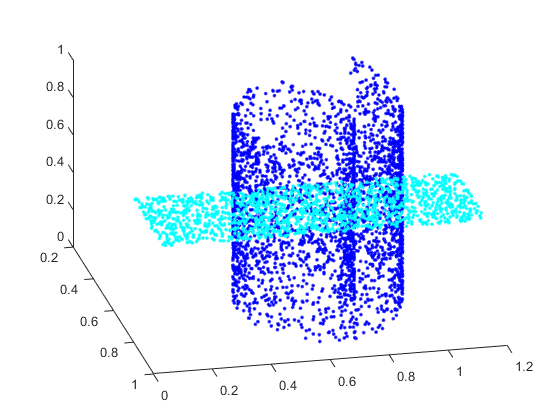}
  \caption{Embedded Data (input) in 3D}
  \label{fig:intersectingb}
\end{subfigure}
\begin{subfigure}{0.3\textwidth}
\includegraphics[height=0.2\textheight]{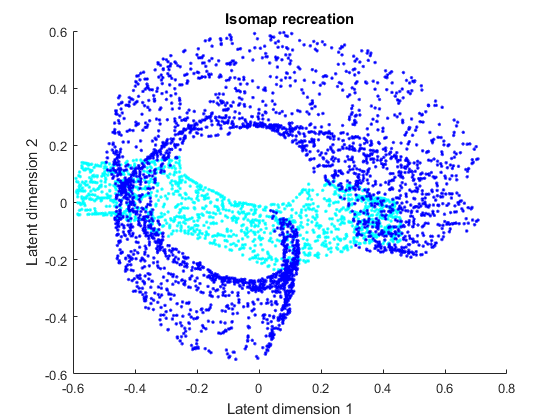} 
  \caption{Isomap/M-Isomap Output}
  \label{fig:intersectingc}
\end{subfigure}
\begin{subfigure}{0.3\textwidth}
\includegraphics[height=0.2\textheight]{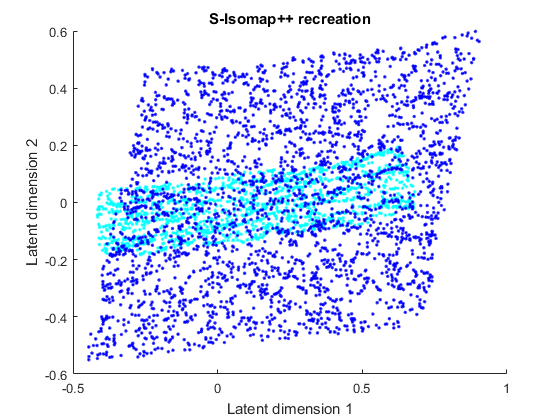}
  \caption{Proposed S-Isomap++ Output}
  \label{fig:intersectingd}
\end{subfigure}
\caption{Multi-manifold {\em intersecting} data set. One set of 2D samples (blue) in (a) are embedded into 3D in (b) via the Euler Isometric mapping technique~\cite{schoeneman2017}. Second set (cyan) are embedded using a linear mapping. The reduction to 2D is obtained using: (c). Isomap/M-Isomap, and (d). the proposed S-Isomap++ algorithm. Both Isomap and M-Isomap give the same output because M-Isomap cannot handle intersecting manifolds and, thus, reverts to a single manifold scenario.}
\label{fig:intersecting}
\end{figure*}

However, the above solution, and other related efforts to adapt NLDR methods to streaming data~\cite{law2006}, rely on the assumption that the data samples lie on a {\em single} low-dimensional manifold. There have been limited attempts that allow for multiple manifolds~\cite{fan2012,fan2016}, however, they assume that the manifolds do not intersect in any ambient space. This is illustrated in Figures~\ref{fig:patches} and~\ref{fig:intersecting}. In Figure~\ref{fig:patches}, the synthetic data set in the top panel consists of four ``patches'' in 2D space which are  embedded onto different regions of a 3D {\em Swiss-Roll}. Thus the 3D patches data set maybe considered as the high-dimensional data set consisting of samples from multiple manifolds. Direct application of Isomap, which assumes that data comes from a single manifold, results in poor recreation of the ground truth (Figure~\ref{fig:patchesc}). An existing method, {\em M-Isomap}~\cite{fan2012}, that explicitly handles multiple manifolds, gives somewhat better results (Figure~\ref{fig:patchesd}). In Figure~\ref{fig:intersecting}, the synthetic data set consists of data from two 2D manifolds embedded in a 3D space, as an isometric swiss-roll and a plane, intersecting with each other. In this case, both Isomap and M-Isomap fail (Figure~\ref{fig:intersectingc}), primarily because M-Isomap assumes that the multiple manifolds do not intersect.

The core contribution of this paper is a streaming non-linear dimensionality reduction algorithm, called {\bf S-Isomap++}. The algorithm assumes that the high dimensional input data consists of samples that truly lie on one or more, potentially intersecting, low-dimensional manifolds and are embedded into the high dimensional space via non-linear transformations. The proposed algorithm extends the widely used Isomap algorithm to handle multiple intersecting manifolds in a streaming setting. Thus, the proposed algorithm operates under one of the least restrictive set of assumptions, explored so far in the context of NLDR methods (See Figures~\ref{fig:patchese} and~\ref{fig:intersectingd}). Moreover, the ability  to handle large streams of data makes it highly applicable in a broad variety of domains. 

Another contribution of the paper is a novel {\em tangent based clustering} strategy to separate samples from the input batch, in the original high-dimensional space, into different clusters. Each cluster is processed independently to obtain the manifold and the corresponding low-dimensional reduction of the corresponding data samples, using Isomap. The reduced data samples are then mapped into a common {\em ambient} space by exploiting the relationship between the samples across the clusters in the original space. The streaming samples are then mapped, in parallel, on each manifold. An evaluation strategy is employed to choose the best manifold for each streaming sample.

The rest of the paper is organized as follows: we provide necessary background about manifold learning in Section~\ref{sec:preliminaries}. Related works are discussed in Section~\ref{sec:related_work}. The proposed algorithm, S-Isomap++, is presented in Section~\ref{sec:methodology}. Experimental results on synthetic and benchmark datasets, are summarized in Section~\ref{sec:results_analysis}.
%
%

%
%


%
%
%


\section{Background and Motivation}
\label{sec:preliminaries}
\begin{figure}
\centering
\includegraphics[width=0.4\textwidth]{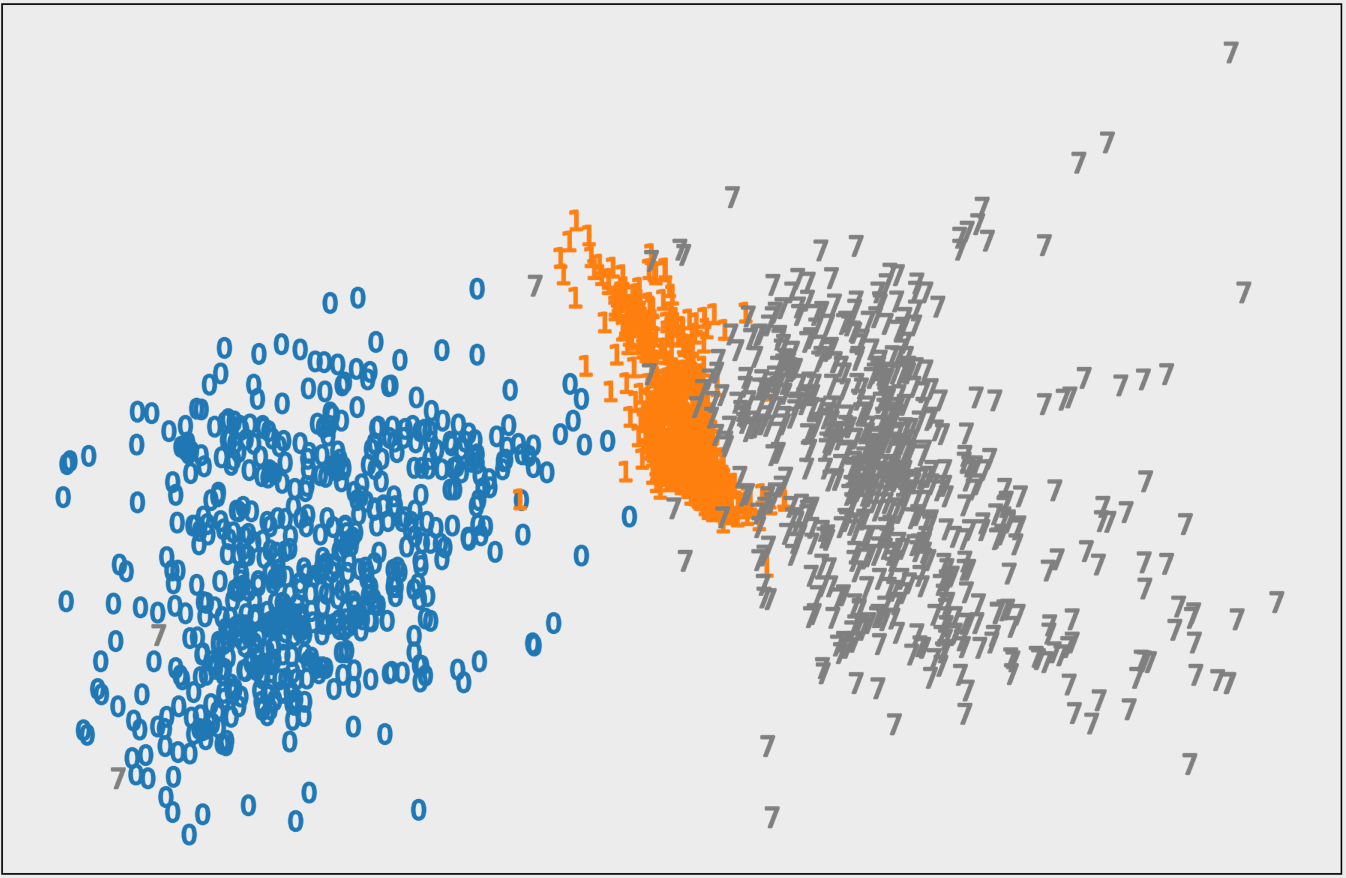}
\caption{2-D reduction of a sample of images from the MNIST digits dataset. Real-world data generally lies near multiple manifolds and is usually separated by regions of low density.}
\label{fig:mnist_data}
\end{figure}

Our motivation for this work stems from one of the foundation principles of {\em Manifold Learning}, which assumes that the distribution of the data in the high-dimensional observed space is not uniform and in reality, the data lies near a non-linear low-dimensional manifold embedded in the high-dimensional space. In many real-world problems such as those resulting from multi-modal or unevenly sampled distributions, the data lies on multiple manifolds of possibly different ``dimensionalities'' and is typically separated by regions of low density as depicted in Figure~\ref{fig:mnist_data}. Thus, to find a representative low-dimensional embedding of the data, one needs to first cluster the data appropriately and subsequently find a low-dimensional representation for the data in each cluster. Even then, manifolds can be very close to each other and can have arbitrary intrinsic dimensions, curvature and sampling which makes it a hard problem to solve.
\subsection{Defining a Manifold}
Mathematically, a manifold $\mathcal{M}$ is defined as a metric space with the following property: if $\mathnormal{x} \in \mathcal{M}$, then there exists some neighborhood $\mathcal{U}$ of $\mathnormal{x}$ and $\exists \mathnormal{n}$ such that $\mathcal{U}$ is homeomorphic to $\mathbb{R}^\mathnormal{n}$~\cite{spivak1979}.

The global structure of the high-dimensional ambient space can be more complicated. Usually manifolds are embedded in high-dimensional spaces, but the intrinsic dimensionality is typically low due to fewer degrees of freedom in the underlying data generating process.

\subsection{Nonlinear Dimensionality Reduction}
\begin{figure*}
  \centering
  \includegraphics[scale=0.7]{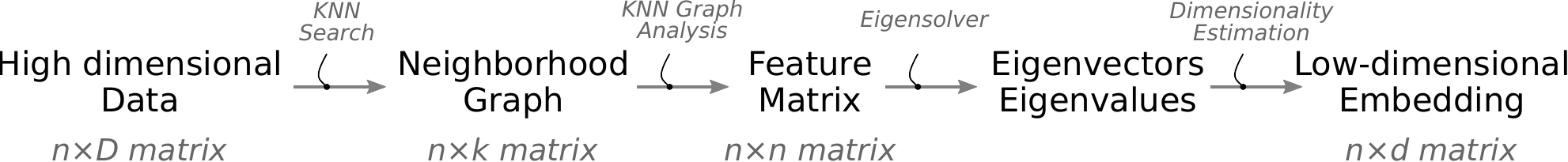}
  \caption{General non-linear spectral dimensionality reduction workflow.}
\label{fig:nlsdr}
\end{figure*}
Typically, nonlinear dimensionality reduction (NLDR) techniques are used as learning methods for discovering the underlying low-dimensional structure from samples from high-dimensional data. Existing techniques typically exploit either the global (Isomap, Minimum Volume Embedding~\cite{weinberger2005}) or local (LLE, Laplacian Eigenmaps~\cite{belkin2002}) properties of the manifold to map each high-dimensional point ${\bf x}_i\in\mathbb{R}^D$ to its corresponding low-dimensional embedding, ${\bf y}_i\in\mathbb{R}^d$. They are used as a generic non-linear, non-parametric technique to approximate probability distributions in high-dimensional spaces. 

The Isomap algorithm, being a global NLDR technique should ideally provide a more faithful representation and preserve geometry irrespective of scale i.e. map data samples which are close in the manifold to points which are close in the low-dimensional embedding and similarly for distant samples. However, it struggles when dealing with multi-modal and non-uniform distributions.


Most existing NLDR techniques, perform a similar series of data transformations as shown in Figure~\ref{fig:nlsdr}. First, a neighborhood graph is constructed, where each node of the graph is connected to its $k$ nearest neighbors. This involves computing $\mathcal{O}(n^2)$ pairwise distance values. Next, a feature matrix is computed from this neighborhood graph, which encodes properties of the data that should be preserved during dimensionality reduction. For example, in the Isomap formulation, the feature matrix stores shortest paths between each pair of points in the neighborhood graph, which is an approximation of the actual geodesic distance between the points. The cost to compute the feature matrix generally varies in the range $\mathcal{O}(n)$ and $\mathcal{O}(n^3)$. To obtain the low representation of the input data, the feature matrix is factorized and the first $d$ eigen vectors/values form the output $Y$. This step has a $\mathcal{O}(n^3)$ cost. 


When used on data streams, NLDR methods typically have to recompute the entire manifold for every new streaming data point, which is computationally expensive. In such scenarios, there is the need for incremental techniques (Out-of-Sample technique~\cite{law2006}, S-Isomap~\cite{schoeneman2017}), which can process the new streaming points ``cheaply'', compared to the traditional batch techniques without affecting the quality of the embedding significantly. 

\subsection{Handling Multiple Manifolds}
In the ideal scenario, when manifolds are densely sampled and sufficiently separated, existing NLDR methods can be extended to perform clustering before applying the dimensionality reduction step~\cite{polito2002, fan2012}, by choosing an appropriate local neighborhood size so as not to include points from other manifolds and still be able to capture the local geometry of the manifold. However, if the manifolds are close or intersecting (See Figures.~\ref{fig:intersecting},~\ref{fig:sr_2p}), such methods typically fail.

\section{Related Work}
\label{sec:related_work}
Most existing NLDR techniques can only deal with a single manifold which leads to them discovering error-prone low dimensional embeddings given inter-manifold distances are usually much larger than the intra-manifold distances. 

Wu \emph{et al.}~\cite{wu2004} was among the earliest attempts to work with multiple manifolds via NLDR techniques. Since then, other sophisticated approaches~\cite{fan2012, hettiarachchi2015, torki2010, yang2011, elhamifar2011} have emerged, apart from techniques in the area of manifold alignment ~\cite{wang2008,ham2005} and manifold clustering~\cite{souvenir2005,fan2012}. Some assumed a supervised setting~\cite{hettiarachchi2015,torki2010}, and learn multiple sub-manifolds corresponding to different given classes in a dataset. The MMDA method~\cite{yang2011} is based on Locality Preserving Projections. Similarly, the SMCE algorithm~\cite{elhamifar2011} makes assumptions about sparsity and linearity of the embedding.

There have been earlier attempts to cluster sub-manifolds~\cite{souvenir2005,fan2012}, which are primarily based on the idea of forming a graph with edges only between a node and its nearest neighbors. However, these methods cannot deal with intersecting manifolds when it is possible for the local neighborhood of a point to have nearest neighbors from different sub-manifolds. Manifold alignment approaches~\cite{wang2008,ham2005} typically align manifolds using a set of correspondences between data points. Whereas~\cite{wang2008} uses Procrustes Analysis,~\cite{ham2005} tries to solve a constrained embedding problem, where the embeddings of the corresponding points from different sets are constrained to be identical. 

%
%
In a batch setting, the M-Isomap~\cite{fan2012} algorithm comes close to our proposed work. The algorithm attempts to work with multiple manifolds embedded in a high-dimensional space. First, it performs clustering to identify the individual sub-manifolds via a nearest neighbor approach and subsequently runs Isomap on each of these sub-manifolds. Finally, it stitches the sub-manifolds together via a set of support points, by finding an optimal transformation between the embeddings uncovered by Multidimensional Scaling(MDS)~\cite{kruskal1978} and Isomap, respectively. However, the nearest neighborhood clustering strategy employed can misrepresent individual sub-manifolds if they are intersecting and/or very close to each other by grouping them together (See Figure~\ref{fig:intersecting}).
%
%
%
%

\section{Methodology}
\label{sec:methodology}
There are two key challenges that a streaming manifold learning algorithm has to address:
\begin{inparaenum}[1)]
\item handle streaming data in a scalable manner, and,
\item learn in presence of multiple, possibly intersecting, manifolds.
\end{inparaenum}

The proposed S-Isomap++ algorithm follows the two-phase strategy proposed in our earlier work~\cite{schoeneman2017}, where we first learn exact manifolds from an initial batch, and then employ a computationally inexpensive mapping method to process the remainder of the stream. An error metric is used to decide on when to {\em switch} from expensive and exact learning to inexpensive and approximate mapping~\cite{schoeneman2017}. To address the second challenge, we first cluster the batch data using a {\em tangent-based manifold clustering} approach and then apply exact Isomap on each cluster. The resulting low-dimensional data for the clusters is then {\em stitched} together to obtain the data reduced to a low (and closer to true) dimensionality. 

The overall S-Isomap++ algorithm is outlined in Algorithm~\ref{alg:sisomap_pp}. The algorithm takes a batch data set, $\mathcal{B}$ and the streaming data, $\mathcal{S}$ as inputs such that, $\mathcal{B},\mathcal{S} \in \mathbb{R}^D$. Note that in practical applications, one might not have data split into batch and streaming parts. In that scenario, one may track the quality of the output of the batch phase using suitable error metrics~\cite{schoeneman2017}, and switch when a reliable solution for the batch is obtained. For simplicity, we will assume that the optimal batch size has been pre-determined. The processing is split into two phases: a batch learning phase (Lines 1--12) and a streaming phase (Lines 13--20). The batch learning phase consists of three steps:

\begin{itemize}
\item Step 1: Cluster samples in $\mathcal{B}$ into $p$ clusters (Line 1).
\item Step 2: Learn $p$ individual manifolds corresponding to each cluster, and map samples within each cluster to a low-dimensional representation\footnote{The true dimensionality of the manifolds corresponding to the clusters can vary. We assume that the true dimensionality for each cluster has been determined using techniques such as studying the spectral properties of the geodesic distance matrix computed as part of Isomap learning (See Figure~\ref{fig:msvd}).} (Lines 6--7).
\item Step 3: Map reduced samples from individual manifolds into a global reduced space (Lines 8--12).
\end{itemize}

In the streaming phase, each sample in the stream set $\mathcal{S}$ is mapped onto each of the $p$ manifolds by using an inexpensive mapping procedure (Lines 14-17). The nearest manifold is identified by comparing each reduced representation of the sample to the ``center'' of each manifold (Line 18), and choosing the corresponding reduced representation for the stream sample (Line 19). 

The individual components of the proposed S-Isomap++ algorithm are discussed in the subsequent subsections.


\begin{algorithm}[h]
  \caption{S-Isomap++}
  \label{alg:sisomap_pp}
  \begin{algorithmic}[1]
    \Require Batch dataset: $\mathcal{B}$, Streaming dataset: $\mathcal{S}$; Parameters: ${\boldsymbol \epsilon}$, ${\boldsymbol k}$, ${\boldsymbol l}$, ${\boldsymbol \lambda}$
    \Ensure  $\mathcal{Y}_\mathcal{S}$: low-dimensional representation for $\mathcal{S}$

    \State $\mathcal{C}_{\mathnormal{i}=1,2 \ldots p}$ $\leftarrow$ \Call{Find\_Clusters}{$\mathcal{B}$, ${\boldsymbol \epsilon}$}
    \State $\xi_{s}$ $\leftarrow$ $\emptyset$
    \Statex
    
    \For{$1\leq i\leq p$}
    \State $\mathcal{LDE}_{i}$ $\leftarrow$ \Call{Isomap}{$\mathcal{C}_{i}$}
    \EndFor
    \Statex
    
    \State $\xi_{s}$ $\leftarrow$ $\bigcup\limits_{\mathnormal{i}=1}^{p} \bigcup\limits_{\mathnormal{j}=\mathnormal{i}+1}^{p}$ $\Call{NN}{\mathcal{C}_{i}, \mathcal{C}_{j}, {\boldsymbol k}}$ $\cup$ \Call{FN}{$\mathcal{C}_{i}, \mathcal{C}_{j}, {\boldsymbol l}}$ 

    \Statex
    \State $\mathcal{GE}_{s}$ $\leftarrow$ \Call{MDS}{$\xi_{s}$}

    \Statex

    \For{$1\leq j\leq p$}
    \State $\mathcal{I}$ $\leftarrow$ $\xi_{s} \cap \mathcal{C}_{j}$
    \State $\mathcal{A}$ $\leftarrow$ \bigg[\begin{tabular}{c}
  $\mathcal{LDE}^\mathcal{I}_{j}$\\
  $e^{T}$\\ \end{tabular}\bigg]
    \State $\mathcal{R}_{i}, t_{i}$ $\leftarrow$ $\mathcal{GE}_{\mathcal{I}, s} \times \mathcal{A}^{T} {\big( \mathcal{A} \mathcal{A}^{T} + {\boldsymbol \lambda} I \big)}^{-1}$
    \EndFor
    \Statex

    \For{$s \in \mathcal{S}$}
    \For{$1\leq i\leq p$}
    \State $y^{i}_{s}$ $\leftarrow$ $\Call{S-Isomap}{s, \mathcal{C}_{i}}$
    \State $\mathcal{GE}^{i}_{s}$ $\leftarrow$ $\mathcal{R}_{i} y^{i}_{s} + t_{i}$
    \EndFor
    
    \Statex
    
    \State index $\leftarrow$ $\argmin_i \left\vert y^{i}_{s} - \mu(\mathcal{C}_{i}, \mathcal{R}_{i}, t_{i}) \right\vert$ 
    \State $\mathcal{Y}_\mathcal{S}$ $\leftarrow$ $\mathcal{Y}_\mathcal{S} \cup y^{index}_{s}$
    \EndFor
    \Statex
    
    \State\Return $\mathcal{Y}_\mathcal{S}$
    \end{algorithmic}
\end{algorithm}

\begin{algorithm}[h]
  \begin{algorithmic}[1]
  \Function{Find\_Clusters}{$\mathcal{B}$, ${\boldsymbol \epsilon}$}
    \State $\mathcal{S}_{\mathnormal{i}=1,2 \ldots n}$ $\leftarrow$ \Call{MSVD}{$\mathcal{B}$}
    \vspace{1mm}
    \State $labels$ $\leftarrow$ $\textbf{0}_{n \times 1}$
    \State $idx$ $\leftarrow$ \textbf{1}
    \vspace{1mm}
    \While{${labels}_{\mathnormal{i}=1,2 \ldots n} \neq \textbf{0} $}
        \State $\mathcal{C}_{idx}$, $labels$ $\leftarrow$ \Call{Cluster}{$\mathcal{B}, \mathcal{S}, labels, idx$, ${\boldsymbol \epsilon}$}
        \State $idx$ $\leftarrow$ $idx$ $+$ \textbf{1}
    \EndWhile
    \vspace{1mm}
    \State\Return $\mathcal{C}_{\mathnormal{i}=1,2 \ldots p}$
\EndFunction
\end{algorithmic}    

\caption{Tangent Manifold Clustering}
\label{alg:tm_clustering}
\end{algorithm}

\subsection{Clustering Multiple Intersecting Manifolds}

\begin{algorithm}[h]
\begin{algorithmic}[1]
\Function{Cluster}{$\mathcal{B}, \mathcal{S}, labels, index$, ${\boldsymbol \epsilon}$}
\State $\mathcal{C}_{index}$ $\leftarrow$ $\emptyset$, $\mathcal{C}_{old}$ $\leftarrow$ $\emptyset$
\State $\mathcal{I}$ $\leftarrow$ \big\{$\mathnormal{i}  \vert  labels_{\mathnormal{i}} = \textbf{0}$\big\}, $idx$ $\sim$ \Call{Random}{$\mathcal{I}$}

\Statex

\State $\mathcal{C}_{index}$ $\leftarrow$ $\mathcal{C}_{index}$ $\cup$ $\mathcal{B}_{idx}$
\State $\mathcal{C}_{old}$ $\leftarrow$ $\mathcal{C}_{old}$ $\cup$ $\mathcal{B}_{idx}$, $labels_{idx}$ $\leftarrow$ $index$
\State $count_{new}$ $\leftarrow$ \textbf{1}, $mode$ = `L1'

\Statex
    
    \While{$count_{new} > \textbf{0}$}
    \State $count_{new}$ $\leftarrow$ \textbf{0}, $\mathcal{C}_{new}$ $\leftarrow$ $\emptyset$

    \Statex
    
    \For{$\forall \mathnormal{i} \in \mathcal{C}_{old}$}
    \State $\mathcal{I}_{knn}$ $\leftarrow$ \Call{KNN}{$\mathcal{B}$, $\mathnormal{i}$}
    \For{$\forall \mathnormal{j} \in \mathcal{I}_{knn}$}
    \If{$labels_{j} = \textbf{0}$}
    \State $sim_{\mathnormal{i},\mathnormal{j}}$ $\leftarrow$ \Call{Sim}{$\mathcal{S}_{\mathnormal{i}}$, $\mathcal{S}_{\mathnormal{j}}$, $mode$}
    \If{$sim_{\mathnormal{i},\mathnormal{j}}$ $\geq$ ${\boldsymbol \epsilon}$}
    \State $\mathcal{C}_{new}$ $\leftarrow$ $\mathcal{C}_{new}$ $\cup$ $\mathcal{B}_{j}$
    \State $labels_{j}$ $\leftarrow$ $index$
    \State $count_{new}$ $\leftarrow$ $count_{new} + \textbf{1}$ 
    \EndIf
    \EndIf
    
    \Statex
    
    \EndFor
    \EndFor
    
    \Statex
    
    \State $\mathcal{C}_{index}$ $\leftarrow$ $\mathcal{C}_{index}$ $\cup$ $\mathcal{C}_{new}$, $\mathcal{C}_{old}$ $\leftarrow$ $\mathcal{C}_{new}$
    
    \EndWhile

    \Statex
    
    \State\Return $\mathcal{C}_{index}, labels$
\EndFunction
\end{algorithmic}
\caption{Incremental Partitioning Strategy}
\label{alg:cluster}
\end{algorithm}

The objective of the first step in Algorithm~\ref{alg:sisomap_pp} is to separate the batch samples into clusters, such that each cluster corresponds to one of the multiple manifolds present in the data. Note that, in this paper, we do not assume that the number of manifolds ($p$) is specified; it is automatically inferred by the clustering algorithm. In cases of uneven/low density sampling, the clustering strategy discussed might possibly generate many small clusters. In such cases, one can try to merge clusters, based on their affinity/closeness to allow the number of clusters to remain within required limits. Given that the batch samples lie on low-dimensional and potentially intersecting manifolds, it is evident that the standard clustering methods, such as K-Means~\cite{jain1999}, that operate on the observed data in $\mathbb{R}^D$, will fail in correctly identifying the clusters.

To handle this challenge, we propose a novel clustering algorithm that is based on the notion of smoothness of manifold surfaces. Consider a single batch data sample, ${\bf x}_i \in \mathbb{R}^D$. Let $\mathcal{N}({\bf x}_i)$ be the set of $k$ nearest neighbor samples of ${\bf x}_i$ in the batch $\mathcal{B}$. Let $\mathcal{T}_i$ denote a $d'$ dimensional {\em tangent plane} represented using $d'$ basis vectors, ${\bf t}_{i1}, {\bf t}_{i2}, \ldots, {\bf t}_{id'}$, i.e., $\mathcal{T}_i = span({\bf t}_{i1}, {\bf t}_{i2}, \ldots, {\bf t}_{id'})$.  Here, $d'$ denotes the intrinsic dimensionality of the tangent plane. We assume that each ${\bf x}_i$ belongs to a single manifold $\mathcal{M}_j, \exists j \in \big\{1,2 \ldots p\big\}$. 

The proposed clustering algorithm (Algorithm~\ref{alg:tm_clustering}) is based on the following intuition: For a given sample, ${\bf x}_i$, and its neighbor ${\bf x}_j \in \mathcal{N}({\bf x}_i)$: 
\begin{align}
\text{If }\mathcal{M}_i = \mathcal{M}_j \Rightarrow \phi(\mathcal{T}_i,\mathcal{T}_j) \ge {\bm \epsilon}
\label{eqn:int1}
\end{align}
$\phi(\mathcal{T}_i,\mathcal{T}_j) = \cos{\theta}$, where $\theta$ is the angle between the two tangent planes\footnote{$\mathcal{T}_i$ and $\mathcal{T}_j$ are the tangent planes for the samples ${\bf x}_i$ and ${\bf x}_j$.}, $\mathcal{T}_i$ and $\mathcal{T}_j$.
Similarly,
\begin{align}
\text{If }\mathcal{M}_i \neq \mathcal{M}_j \Rightarrow \phi(\mathcal{T}_i,\mathcal{T}_j) < {\bm \epsilon}
\label{eqn:int2}
\end{align}
In other words, within a tight neighborhood, a given data sample and its neighbors are expected to lie on tangent planes that are approximately similar in orientation, and, thus, the cosine of the angle between the two planes will be closer to 1 ($\cos{\theta} \approx 1$). However, if a sample's neighborhood contains samples that lie on other intersecting manifolds, their tangent planes should be significantly different, and $\cos{\theta} \ll 1$.
\subsubsection{Learning a Tangent Plane for a Given Sample}
\label{subsubsection:msvd}
We use {\em Multiscale Singular Value Decomposition} (or MSVD~\cite{little2009}) on the local neighborhood of ${\bf x}_i$, to determine basis vectors, ${\bf t}_{i1}, {\bf t}_{i2}, \ldots, {\bf t}_{id'}$, which define the tangent plane, $\mathcal{T}_i$.
Use of SVD allows us to follow the intuitions expressed in~\eqref{eqn:int1} and~\eqref{eqn:int2}, since it explores directions in which the spread of points is maximal. In the presence of multiple intersecting manifolds, these directions get mangled up, whereas non-intersecting regions have better agreement with regards to principal directions. 
%


MSVD allows us to deal with the problem of estimating the intrinsic dimension of noisy, high-dimensional point clouds. For the linear case, SVD analysis can estimate the intrinsic dimensionality of $\mathcal{M}$ correctly, with high probability. However, when $\mathcal{M}$ is a nonlinear manifold, curvature forces the dimensionality of the best-approximating hyperplane to be much higher, which hinders attempts to uncover the true intrinsic dimensionality of $\mathcal{M}$.

\begin{figure}[p]
\centering
\includegraphics[scale=0.7]{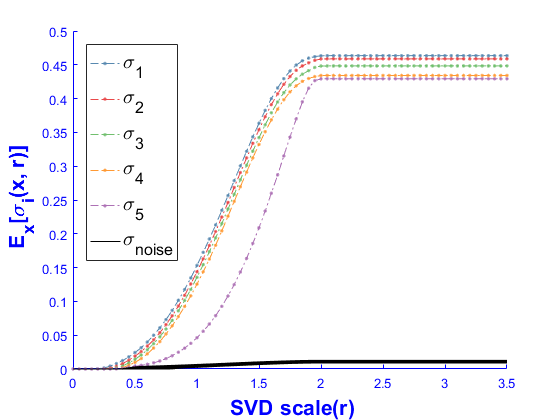}
\caption{Multiscale SVD on a noisy $\mathbb{R}^{5}$ sphere embedded in $\mathbb{R}^{100}$ ambient space.}
\label{fig:msvd}
\end{figure}

MSVD estimates the intrinsic dimensionality of $\mathcal{M}$ by computing the singular values, $\sigma_{i}^{z, r}$ for all $\forall z\in \mathcal{M}$ at different scales $r > 0$ and $i\in \{1,2, \ldots D \}$. Small values of $r$ lead to not enough samples in $\mathcal{B}(z,r)$, while large values of $r$ lead to curvature making the SVD computation over estimate the intrinsic dimensionality. At the right scale (value of $r$), the true $\sigma_{i}^{z, r}$'s separate from the noise $\sigma_{i}^{z, r}$'s due to their different rates of growth and the true dimensionality of $\mathcal{M}$ is revealed. Figure~\ref{fig:msvd} demonstrates how $\sigma_{i}^{z, r}$ behave over different scales when MSVD is done a noisy $\mathbb{R}^{5}$ sphere embedded in $\mathbb{R}^{100}$ ambient space. Notice how the noise dimensions decay out, leaving only the primary components at the appropriate scale.

\subsubsection{Computing Angle Between Two Tangent Planes}
\label{subsubsection:sim}
We explore several strategies of computing the similarity between a pair of tangent planes, $\mathcal{T}_i$ and $\mathcal{T}_j$. As mentioned earlier, this is equivalent to computing the cosine of the angle between the two planes. We consider one approach, as proposed by Gunawan \emph{et al.} \cite{gunawan2005}. Let $\mathcal{T}_i$ and $\mathcal{T}_j$ be orthonormal subspaces\footnote{One can use QR factorization to orthonormalize any subspace, which is not already orthonormal.}. If $\theta$ is the angle between $\mathcal{T}_i$ and $\mathcal{T}_j$, then:
\begin{equation}
\phi(\mathcal{T}_i,\mathcal{T}_j) = \cos{\theta} = \sqrt{det(\mathcal{N}\mathcal{N}^\top)}
\label{eqn:gunawan}
\end{equation}
where $\mathcal{N}$ is a matrix, such that $\mathcal{N}[u][v] = \langle {\bf t}_{iu},{\bf t}_{jv}\rangle$, where ${\bf t}_{iu}$ is the $u^{th}$ basis vector for $\mathcal{T}_i$ and ${\bf t}_{jv}$ is the $v^{th}$ basis vector for $\mathcal{T}_j$. 
Additionally, when the dimensionality of $\mathcal{T}_i$ and $\mathcal{T}_j$ is same, the expression simplifies to:
\begin{equation}
\phi(\mathcal{T}_i,\mathcal{T}_j) = \cos{\theta} = \left\vert det(\mathcal{N})\right\vert
\label{eqn:gunawansimple}
\end{equation}
Alternately, one can use the following procedure. Without loss of generality, let us assume that ${\bf t}_{i1}, {\bf t}_{i2}, \ldots, {\bf t}_{ik}$ are the singular vectors for the plane $\mathcal{T}_i$ corresponding to the top $k$ singular values. Similarly, let ${\bf t}_{j1}, {\bf t}_{j2}, \ldots, {\bf t}_{jk}$ be the top-$k$ singular vectors for the plane $\mathcal{T}_j$. Then we can compute $\phi(\mathcal{T}_i,\mathcal{T}_j)$ as:
\begin{equation}
\phi(\mathcal{T}_i,\mathcal{T}_j) = \frac{1}{k} \sum_{l=1}^k\vert{\bf t}_{il}^\top{\bf t}_{jl}\vert
\label{eqn:l1}
\end{equation}
We refer to the above as the $L1$ metric. In the same way, one can define the $L2$ metric as:
\begin{equation}
\phi(\mathcal{T}_i,\mathcal{T}_j) = \sqrt{\frac{1}{k} \sum_{l=1}^k({\bf t}_{il}^\top{\bf t}_{jl})^2}
\label{eqn:l2}
\end{equation}




%
%
%
%

\subsubsection{Tangent Manifold Clustering Algorithm}
%
%
The proposed tangent manifold clustering strategy is outlined in Algorithm~\ref{alg:tm_clustering}. Algorithm~\ref{alg:cluster} is the support method to the above. The inputs to the Algorithm~\ref{alg:tm_clustering} are the batch dataset $\mathcal{B}$ and a threshold value ${\bm \epsilon}$. 

Algorithm~\ref{alg:tm_clustering} initially calls \Call{MSVD}{$\cdot$} (See Section~\ref{subsubsection:msvd}) on the input batch set, $\mathcal{B}$, to decide on an appropriate scale $r$ to use and subsequently to extract the top-$k$ singular vectors $\mathcal{S}_{\mathnormal{i}=1,2 \ldots n}$ for all ${\bf x}_i\in\mathcal{B}$, at the scale $r$. Initially all points are unlabeled i.e. {\em labels} is all zeros initially. Algorithm~\ref{alg:tm_clustering} calls \Call{Cluster}{$\cdot$} repeatedly till all ${\bf x}_i\in\mathcal{B}$ have labels assigned to them, which represents the different clusters, $\mathcal{C}_{\mathnormal{i}}$ for $\mathnormal{i} = 1,2 \ldots m$ where $\bigcup\limits_{\mathnormal{i}=1}^{m} \mathcal{C}_{i} = \mathcal{B}$. 
%

Algorithm~\ref{alg:cluster}, which contains the function \Call{Cluster}{$\cdot$}\footnote{We use `L1' as the mode by default (Line 6) since it provides the best accuracy.}, works as follows: it picks a currently unlabeled ${\bf x}_i$ at random, and assigns it to a new cluster $\mathcal{C}_{index}$. Subsequently, it looks at the unassigned nearest neighbors of ${\bf x}_i$ i.e. ${\bf x}_j\in\mathcal{N}({\bf x}_i)$ and checks to see how close their tangent planes are. If they are similar enough i.e. the similarity score $\phi(\mathcal{T}_i, \mathcal{T}_j) \geq \epsilon$, then the unassigned nearest neighbor is assigned to $\mathcal{C}_{index}$. The algorithm proceeds similarly in a breadth-first manner till no new points remain to be tested.


It internally calls Algorithm \Call{Sim}{$\cdot$} to measure similarity, using one of the three strategies, discussed in Section~\ref{subsubsection:sim} (See~\eqref{eqn:gunawansimple},~\eqref{eqn:l1}, and~\eqref{eqn:l2}).

\begin{algorithm}[p]
\begin{algorithmic}[1]
\Function{Sim}{$\mathcal{S}_i$, $\mathcal{S}_j$, $mode$}

\State $\eta_{\mathnormal{i}=1,2 \ldots k}$ $\leftarrow$ extract($\mathcal{S}_i$)

\State $\kappa_{\mathnormal{i}=1,2 \ldots k}$ $\leftarrow$ extract($\mathcal{S}_j$)
\State
\If{$mode = `L1'$}
\State $score$ $\leftarrow$ $\frac{1}{k} \sum_{i=1}^{k} \left\vert {\eta_{i}}^{T}\kappa_{i}\right\vert$
\State
\ElsIf{$mode = `L2'$}
\State
\State $score$ $\leftarrow$ $\sqrt{\sum_{i=1}^{k} \frac{1}{k} ({{\eta_{i}}^{T}\kappa_{i}})^{2}}$
\State
\ElsIf{$mode = `HG'$}
\State $\mathcal{M}_{\eta}$ $\leftarrow$ $matrix(\eta_{\mathnormal{i}=1,2 \ldots k})$
\State $\mathcal{M}_{\kappa}$ $\leftarrow$ $matrix(\kappa_{\mathnormal{i}=1,2 \ldots k})$
\State
\State $\mathcal{M}$ $\leftarrow$ ${\mathcal{M}_{\eta}}^{T}\mathcal{M}_{\kappa}$
\State $score$ $\leftarrow$ $\left\vert det(\mathcal{M})\right\vert$
\EndIf

\State
\State\Return $score$
\EndFunction
\end{algorithmic}
\caption{Similarity between tangent planes between points}

\label{alg:sim}
\end{algorithm}

\subsection{Processing multiple manifolds}
The S-Isomap++ algorithm independently learns the manifolds for each cluster (Lines 3--5). However, since these manifolds are not necessarily aligned with respect to each other, an additional step is needed to represent the reduced samples from each cluster into a common space. We refer to this process as {\em stitching}, and is essential to recreate the final reduced representation. This step, similar to the approach in M-Isomap, maintains the information of the global location of different manifolds using a set of support points which form the skeleton on which it can later places the different manifolds. This support set is formed using the $k$ nearest neighbor pairs as well as the $l$ farthest neighbor pairs between every pair of manifolds present i.e. $\forall \{\mathcal{C}_i, \mathcal{C}_j\}_{j \neq i}$, let $\mathcal{X}_{i,j} \in \mathbb{R}^{\left\vert \mathcal{C}_i \right\vert \times \left\vert \mathcal{C}_j \right\vert}$ denote the $\mathbb{R}^{D}$ Euclidean distance matrix between all points in clusters $\mathcal{C}_i$ and $\mathcal{C}_j$, then support set $\xi_{s}$ contains the co-ordinates (index sets $\mathcal{I}_{i}$ and $\mathcal{I}_{j}$ from $\mathcal{C}_i$ and $\mathcal{C}_j$ respectively) of both the smallest $k$ values as well as the largest $l$ values in $\mathcal{X}_{i,j}$. The former are calculated by method \Call{NN}{$\cdot$} and the latter \Call{FN}{$\cdot$} (Line 6). Subsequently, a global reduced space embedding $\mathcal{GE}_s$ for this support set is calculated using MDS (Line 7). After this, for each manifold $\mathcal{M}_j, \exists j \in \big\{1,2 \ldots p\big\}$, a least-squares problem is solved to generate the transformation components $\mathcal{R}_{i}, t_{i}$ which can project reduced samples from each cluster into the global space (Lines 8--12).

\subsection{Mapping Streaming Samples}
In the streaming part, each sample in the stream set $\mathcal{S}$ is mapped onto each of the $p$ manifolds in parallel, using the inexpensive \Call{S-Isomap}{$\cdot$} algorithm proposed in our earlier work~\cite{schoeneman2017} (Line 15) and subsequently mapped to the global space using $\{\mathcal{R}_{i}, t_{i}\} \,\exists i \in \big\{1,2 \ldots p\big\}$ (Line 16). The nearest manifold is identified by comparing each reduced representation of the sample to the mean \Call{$\mu$}{$\cdot$} of each manifold (Line 18), and choosing the corresponding reduced representation for the stream sample (Line 19).

\section{Results and Analysis}
\label{sec:results_analysis}
\subsection{Experimental Setup}
We present several experiments here on a variety of data sets to illustrate the behavior of different approaches proposed in the Section \ref{sec:methodology}. 

We use four different datasets in our experiments. Given swiss roll datasets are typically used for evaluating manifold learning algorithms, we use the Euler Isometric Swiss Roll dataset, proposed by Schoeneman \emph{et al.}~\cite{schoeneman2017}, wherein a $\mathbb{R}^{2}$ data set having $n=3000$ points, chosen at random, are embedded into $\mathbb{R}^{3}$ using a non-linear function $\psi(\cdot)$. We use this in conjunction with a $\mathbb{R}^{3}$-dimensional hyperplane passing through it as shown in Figure.~\ref{fig:intersectingb} having $n=1500$ points, chosen at random. We know the ground truth for both parts (See Figure.~\ref{fig:intersectinga}). We use this to evaluate the S-Isomap++ algorithm as shown in Figure.~\ref{fig:intersecting}. We also use an extension of this, wherein two $\mathbb{R}^{3}$-dimensional hyperplanes pass through the Isometric Swiss Roll, wherein the points are chosen in random and each hyperplane has $n=3000$ points, as shown in Figure.~\ref{fig:sr_2p}.

Apart from this, we use different artificial datasets consisting of intersecting manifolds i.e. two intersecting $\mathbb{R}^{3}$-dimensional unit hyperspheres, having $n=1000$ points each and a $\mathbb{R}^{3}$-dimensional plane intersecting a $\mathbb{R}^{3}$-dimensional hypersphere, again having $n=1000$ points each, as shown in Figure.~\ref{fig:m_manifold}. We use these datasets to test our tangent manifold approach more rigorously. We also use patches on the Euler Isometric Swiss Roll dataset (Figure.~\ref{fig:patches}) which are Gaussian in nature, to study the effect of the different parameters, apart from evaluating our algorithm, as well as the MNIST digits dataset.

Our evaluation metrics for the experiments primarily focus on \begin{inparaenum}[1)]
\item ability on our tangent manifold clustering strategy to be able to cluster points from multiple intersecting/non-intersecting manifolds correctly,
\item test the quality of the embedding uncovered by our algorithm, for the streaming dataset $\mathcal{S}$, with regards to agreeability with ground truth via an appropriate distance metric, as well as, tightness of clustering and last but not the least, \item scalability of our algorithm over different sizes of both batch and streaming datasets $\mathcal{B}$ and $\mathcal{S}$ respectively.
\end{inparaenum}



\subsection{Results on Artificial Datasets}
\subsubsection{Gaussian patches on Isometric Swiss Roll}
Figures.~\ref{fig:patchesc}, \ref{fig:patchesd}, \ref{fig:patchese} demonstrate the results with this dataset for Isomap, M-Isomap and our approach respectively. Both the M-Isomap and S-Isomap++ algorithms can deal with individual manifolds better than Isomap, which severely deforms the individual clusters. It should also be noted that whereas both the M-Isomap and S-Isomap++ algorithms required small values of $k$ i.e. $k=8$ to operate, Isomap needed values of $k \geq 500$ to even work. As a consequence, idiosyncrasies i.e. short-circuiting become a factor to distort the uncovered embedding. M-Isomap has scaling issues and can only seem to attempt to position the individual manifolds in the global ambient space correctly, without being able to recreate the spread, which defined the individual manifolds. We think that M-Isomap internally normalizes individual manifolds which results in this behavior. Our approach, S-Isomap++ is the most robust in its recreation of the ground truth.
 

\subsubsection{Intersecting Swiss-roll with $\mathbb{R}^{3}$-dimensional plane}
\begin{figure}
\centering
\includegraphics[scale=0.25]{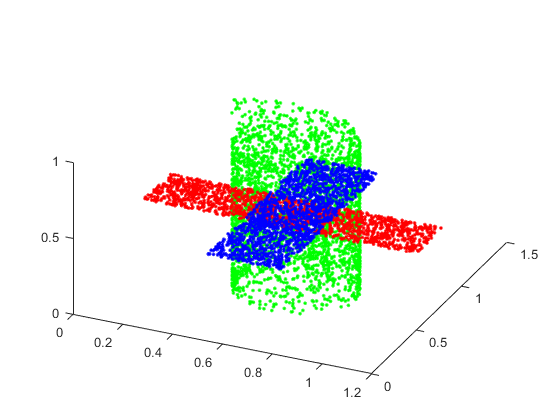}
\medskip
\includegraphics[scale=0.25]{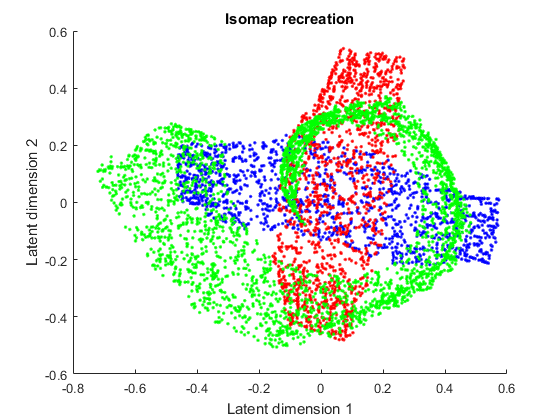}
\includegraphics[scale=0.25]{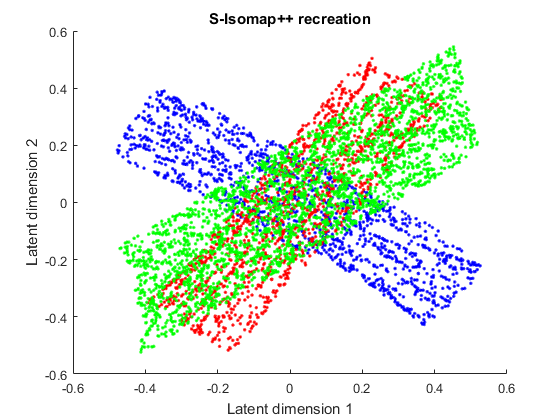}
%
\caption{Top Left: Actual manifolds in $\mathbb{R}^{3}$ space, clustered to demonstrate individual manifolds, Top Right: Recreation by Isomap/M-Isomap, Bottom Row: Recreation by our approach, S-Isomap++.}
\label{fig:sr_2p}
\end{figure}

Figure.~\ref{fig:intersecting} demonstrates our experiments with this dataset. We evaluate different algorithms to see how well they recreate the ground truth (Figure.~\ref{fig:intersectinga}). Both Isomap and M-Isomap produce the same output, given M-Isomap employs a nearest-neighbor based clustering strategy to disambiguate between manifolds, and hence is unable to handle intersecting manifolds, which results in highly distorted recreations of the ground truth. As before, S-Isomap++ produces the most robust recreation of the ground truth. Figure~\ref{fig:sr_2p}, demonstrates how well S-Isomap++ recreates the original manifolds, in case the batch $\mathcal{B}$ is clustered correctly. M-Isomap/Isomap are unable to recreate the ground truth and severely contort the ground truth.


\subsubsection{Tangent Manifold Clustering}
\begin{figure}
\centering
\includegraphics[scale=0.25]{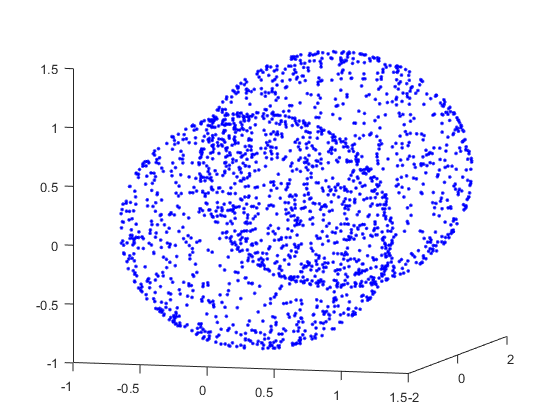}
\includegraphics[scale=0.25]{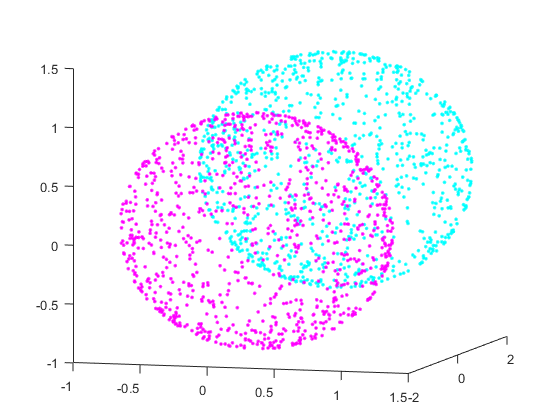}
\medskip
\includegraphics[scale=0.25]{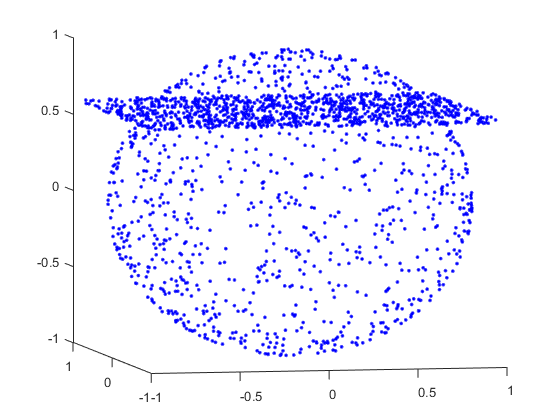}
\includegraphics[scale=0.25]{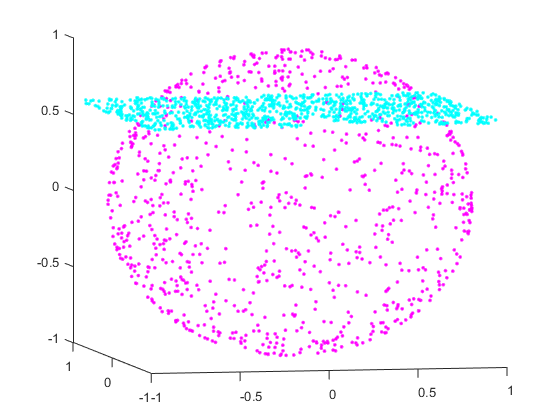}
\caption{Left: Original datasets unclustered, Right: Clustered using the proposed tangent clustering method.}
\label{fig:m_manifold}
\end{figure}

Here we present clustering results for intersecting manifolds. (See Figures.~\ref{fig:intersecting},~\ref{fig:m_manifold} for the different datasets). Table~\ref{table:clustering} below demonstrates accuracy values\footnote{Gunawan's approach was unable to distinguish between the intersecting manifolds scenarios and always clustered them as one and hence its accuracy was $0.5$ in all cases.} with which the L-1, L-2 metric schemes proposed in this work, along with the technique proposed by Gunawan \emph{et al.}~\cite{gunawan2005} clustered the different intersecting manifolds. The L-2 metric performed much better than Gunawan's approach, however the L-1 metric performed the best. The accuracy values are also indicative of the level of difficulty associated with clustering the different scenarios correctly.

\begin{table}[h!]
\centering
\begin{tabular}{| c | c | c | c |} 
 \hline
 \emph{Method} & \emph{L-1} & \emph{L-2} & \emph{Gunawan} \\ 
 \hline
 Sphere-Sphere & \textbf{0.825} & 0.619 & 0.5 \\ 
 \hline
 Sphere-Plane & \textbf{0.759} & 0.602 & 0.5 \\
 \hline
 Swiss Roll-Plane & \textbf{0.838} & 0.621 & 0.5 \\ [1ex]
 \hline
\end{tabular}
\caption{Accuracy scores for the different tangent manifold clustering approaches.}\label{table:clustering}
\end{table}

\subsubsection{Effect of different parameters}
\begin{figure}
\centering
\includegraphics[scale=0.25]{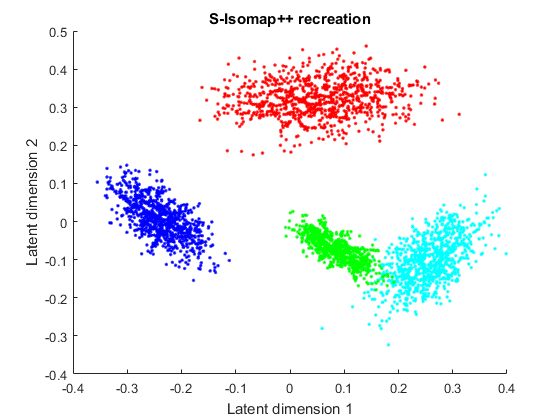}
\includegraphics[scale=0.25]{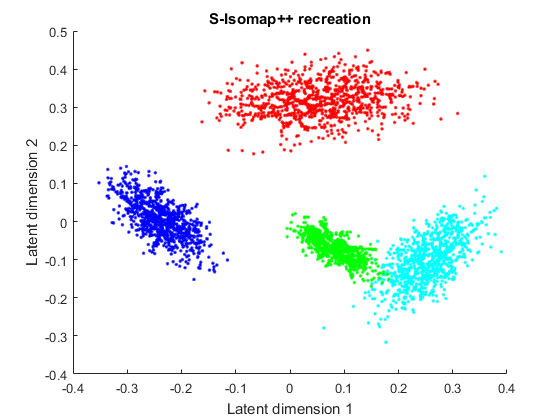}
\medskip
\includegraphics[scale=0.25]{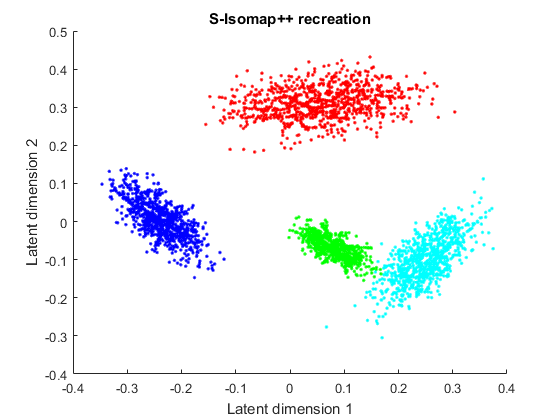}
\includegraphics[scale=0.25]{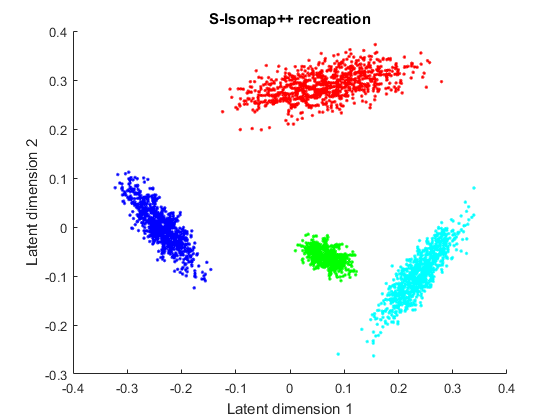}
\caption{Effect of changing $\lambda$. Top Left: $\lambda = 0.01$, Top Right: $\lambda = 0.02$, Bottom Left: $\lambda = 0.04$, Bottom Right: $\lambda = 0.16$}
\label{fig:lambda}
\end{figure}

\begin{figure}
\centering
\includegraphics[scale=0.25]{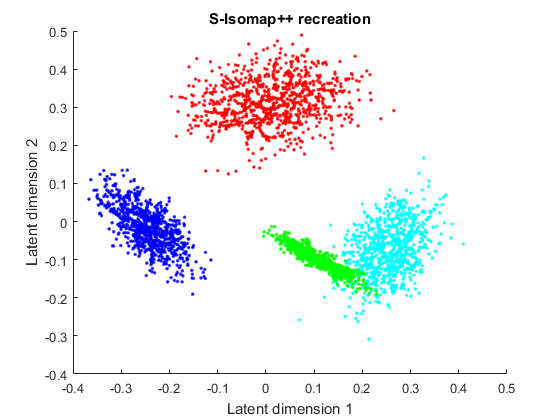}
\includegraphics[scale=0.25]{img/k16_l1}
\medskip
\includegraphics[scale=0.25]{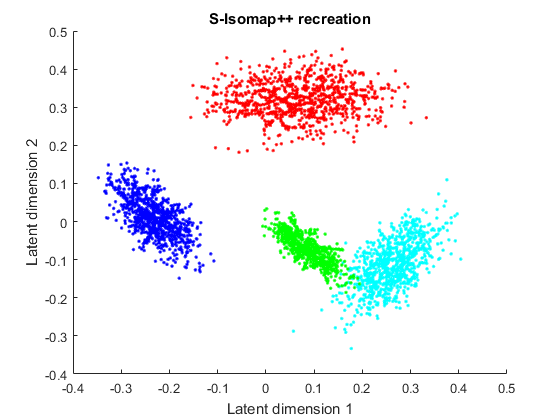}
\includegraphics[scale=0.25]{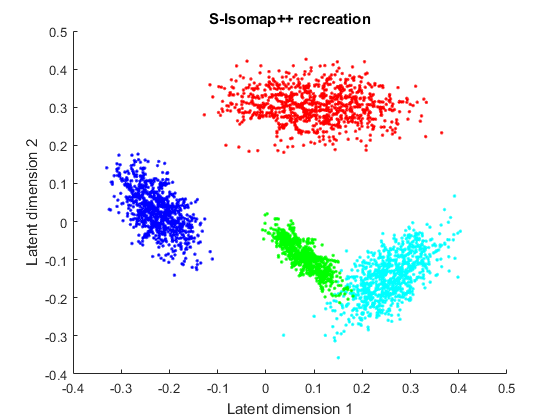}
\caption{Effect of changing $k$. Top Left: $k = 8$, Top Right: $k = 16$, Bottom Left: $k = 24$, Bottom Right: $k = 32$}
\label{fig:k}
\end{figure}

\begin{figure}
\centering
\includegraphics[scale=0.25]{img/k16_l1}
\includegraphics[scale=0.25]{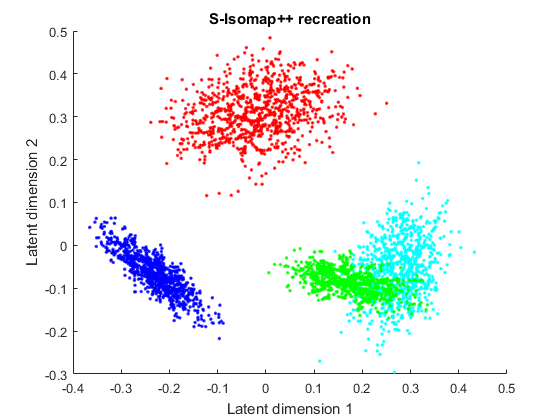}
\caption{Effect of changing $l$. Left: $l = 1$, Right: $l = 4$}
\label{fig:l}
\end{figure}

Here we present results of the effect of changing the different parameters of the S-Isomap++ algorithm, \emph{while keeping all other parameters fixed}. Figures~\ref{fig:lambda},~\ref{fig:k},~\ref{fig:l}, demonstrates the effect of parameter $\lambda$, $k$ and $l$ on the embeddings uncovered by the S-Isomap++ algorithm. Larger values of $k$ seems to make the manifolds more uniform or rounded. Larger values of  parameter $l$ seem to stretch the manifolds. Parameter $\lambda$ seem to separate the manifolds apart when it has larger values. This is really interesting since it means we can use it to visualize manifolds better on account of separability.

\begin{figure}
\centering
\includegraphics[scale=0.45]{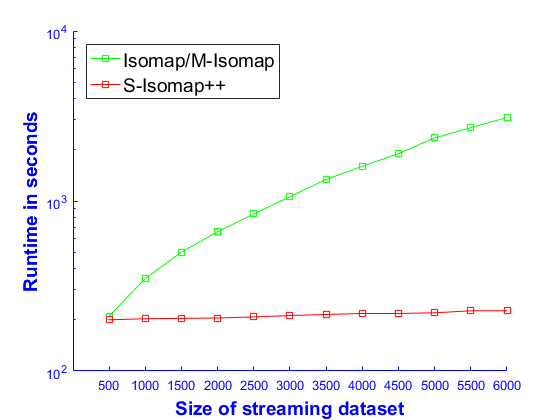}
\caption{The results are in log scale and demonstrate the scalability of our proposed algorithm.}
\label{fig:scalability}
\end{figure}

Figure~\ref{fig:scalability} demonstrates the scalability of our algorithm with regards to streaming data $\mathcal{S}$. Batch $\mathcal{B}$ having size $n=2000$ was used for this experiment. The timing results are in log scale and clearly demonstrate the efficiency gained. M-Isomap has the same result as Isomap since it cannot distinguish between intersecting manifolds and treats them as one. While the run-time for Isomap/M-Isomap increases rapidly with increasing stream size, the run time for S-Isomap++ does not grow much at all, making it highly conducive to large stream processing.

\subsection{Results on MNIST Dataset}
Table~\ref{table:mnist} below shows results for different digits of the MNIST dataset. Using a batch dataset $\mathcal{B}$ of size $n=2000$, a streaming dataset $\mathcal{S}$ of size $m=4000$ was recreated in 3D by the S-Isomap++ algorithm, for each of the digits. Subsequently the 3D recreation was compared to the 3D ground truth obtained by running Isomap on all digits, using the Procrustes Error metric to measure the quality of the recreation.

The Procrustes Error metric determines an optimal alignment between two matrices $\mathcal{X}$ and $\mathcal{Y}$ and returns a goodness-of-fit criterion, based on sum of squared errors. As the results below demonstrate, the recreation error is pretty low, even after embedding in the common global space. This shows the efficacy of the S-Isomap++ algorithm.

\begin{table}[h!]
\centering
\begin{tabular}{| c | c || c | c || c | c |}
\hline
\emph{digit `0'} & \textbf{0.0296} & \emph{digit `3'} & \textbf{0.0364} & \emph{digit `6'} & \textbf{0.0476}\\
\hline
\emph{digit `1'} & \textbf{0.0806} & \emph{digit `4'} & \textbf{0.0586} & \emph{digit `8'} & \textbf{0.0712}\\
\hline
\emph{digit `2'} & \textbf{0.0499} & \emph{digit `5'} & \textbf{0.0449} & \emph{digit `9'} & \textbf{0.0498}\\
\hline
\end{tabular}
\caption{Procrustes error values for different digits of the MNIST dataset, computed by comparing the original with 3D recreation via S-Isomap++.}\label{table:mnist}
\end{table}


%
%

\section{Conclusion}
\label{sec:conclusion}
The proposed S-Isomap++ algorithm allows for scalable non-linear dimensionality reduction of streaming high-dimensional data. By allowing for the samples to belong to multiple manifolds, or sampled non-uniformly from a single manifold, we have developed an algorithm that can be applied to a wide variety of practical settings.  Moreover, the two-phase strategy for streaming Isomap, first proposed in~\cite{schoeneman2017}, and adapted here for multiple manifold learning, allows us to scale a computationally intensive algorithm (Isomap) to arbitrarily large streams.

The ability to cluster data lying on multiple intersecting manifolds is a key innovation, proposed as the Tangent Manifold Clustering algorithm, allows us to automatically identify the number of underlying manifolds. One limitation of the method, however, is that it assumes that all manifolds are represented in the batch data set, which means that a novel manifold behavior that might appear subsequently in the stream, will not be learned. This issue will be studied in future research.

\section{Acknowledgment}
\label{sec:acknowledgement}
This material is based in part upon work supported by the National Science Foundation under award numbers CNS - 1409551 and IIS - 1651475. Computing support was provided by the Center for Computational Research.

\end{document}